\documentclass{article}

    \PassOptionsToPackage{numbers, compress}{natbib}

\usepackage[final]{neurips_2022}

\usepackage[utf8]{inputenc} %
\usepackage[T1]{fontenc}    %
\usepackage{hyperref}       %
\usepackage{url}            %
\usepackage{booktabs}       %
\usepackage{amsfonts}       %
\usepackage{nicefrac}       %
\usepackage{microtype}      %
\usepackage{xcolor}         %
\usepackage{wrapfig}

\usepackage{microtype}
\usepackage{graphicx}
\usepackage{subfigure}
\usepackage{booktabs} %
\usepackage{multirow}
\usepackage[linecolor=red,backgroundcolor=red!25,bordercolor=red,textsize=tiny]{todonotes}

\usepackage{pifont}
\newcommand{\cmark}{\ding{51}}
\newcommand{\xmark}{\ding{55}}
\usepackage{enumitem}
\usepackage{times}

\usepackage{fancyhdr}
\usepackage{xcolor} %
\usepackage{algorithm}
\usepackage{algorithmic}
\usepackage{natbib}
\usepackage{eso-pic} %
\usepackage{forloop}
\usepackage{url}

\usepackage{hyperref}

\newcommand\hmm[1]{\ifnum\ifhmode\spacefactor\else2000\fi>1000 \uppercase{#1}\else#1\fi}

\newcommand{\shp}{\hmm{S}hrink \& Perturb} %
\newcommand{\lwr}{\hmm{L}ayer-wise Re-initialization}
\newcommand{\reinit}{\hmm{r}e-initialization} %
\newcommand{\numexp}{15,000}

\newcommand{\paramdim}{p}
\newcommand{\param}{{\boldsymbol{\theta}}}
\newcommand{\newinitparam}{\param_\text{init}}
\newcommand{\reinitparam}{\param_\text{RI}}
\newcommand{\nn}{f}
\newcommand{\R}{\mathbb{R}}
\newcommand{\loss}{L}
\newcommand{\initdist}{p_\text{init}}
\newcommand{\ritransform}{R}
\newcommand{\numepochs}{N}
\newcommand{\numstages}{T}
\newcommand{\stageindex}{t}
\newcommand{\shrink}{\lambda}
\newcommand{\perturb}{\gamma}
\newcommand{\mask}{{\boldsymbol{m}}}
\newcommand{\preresnet}{PreAct-ResNet-18}
\newcommand{\tin}{Tiny ImageNet}
\newcommand{\numblocks}{K} 
\newcommand{\lwriters}{M}
\newcommand{\lnfrac}{q}
\newcommand{\distillcoeff}{{\beta_\text{distill}}}
\newcommand{\epochsperstage}{P}

\newcommand{\setA}{{\textsc{$\varnothing$}}}
\newcommand{\setB}{{\textsc{d}}}
\newcommand{\setC}{{\textsc{dc}}}
\newcommand{\setD}{{\textsc{dcw}}}

\usepackage{amsmath}
\usepackage{amssymb}
\usepackage{mathtools}
\usepackage{amsthm}

\DeclarePairedDelimiterX{\infdivx}[2]{(}{)}{%
  #1\;\delimsize\|\;#2%
}
\newcommand{\infdiv}{\text{KL}\infdivx}

\usepackage[capitalize,noabbrev]{cleveref}

\newcommand{\LineComment}[1]{\hfill\textit{#1}}
\theoremstyle{plain}

\theoremstyle{definition}

\theoremstyle{remark}

\usepackage[textsize=tiny]{todonotes}
\title{\bf{When Does Re-initialization Work?}}

\author{Sheheryar Zaidi$^1$$^*$\hspace{3mm} Tudor Berariu$^2$$^*$ \hspace{3mm} Hyunjik Kim$^3$ \vspace{1mm} \\ \textbf{Jörg Bornschein$^3$\hspace{3mm} Claudia Clopath$^{2,3}$\hspace{3mm} Yee Whye Teh$^{1,3}$ \hspace{3mm} \textbf{Razvan Pascanu$^3$}} \vspace{1mm} \\
 $^1$University of Oxford,
 $^2$Imperial College London,
 $^3$DeepMind \vspace{1mm} \\ 
 $^*$Equal contribution.\\
 Correspondence to: \texttt{szaidi@stats.ox.ac.uk, t.berariu19@imperial.ac.uk}\\
}

\begin{document}

\maketitle

\begin{abstract}
Re-initializing a neural network during training has been observed to improve generalization in recent works. Yet it is neither widely adopted in deep learning practice nor is it often used in state-of-the-art training protocols. This raises the question of when re-initialization works, and whether it should be used together with regularization techniques such as data augmentation, weight decay and learning rate schedules. In this work, we conduct an extensive empirical comparison of standard training with a selection of re-initialization methods to answer this question, training over 15,000 models on a variety of image classification benchmarks. We first establish that such methods are consistently beneficial for generalization in the absence of any other regularization. However, when deployed alongside other carefully tuned regularization techniques, re-initialization methods offer little to no added benefit for generalization, although optimal generalization performance becomes less sensitive to the choice of learning rate and weight decay hyperparameters. To investigate the impact of re-initialization methods on noisy data, we also consider learning under label noise. Surprisingly, in this case, re-initialization significantly improves upon standard training, even in the presence of other carefully tuned regularization techniques.
\end{abstract}

\section{Introduction}

\begin{figure}[t!]
    \centering
    \includegraphics[width=0.48\linewidth]{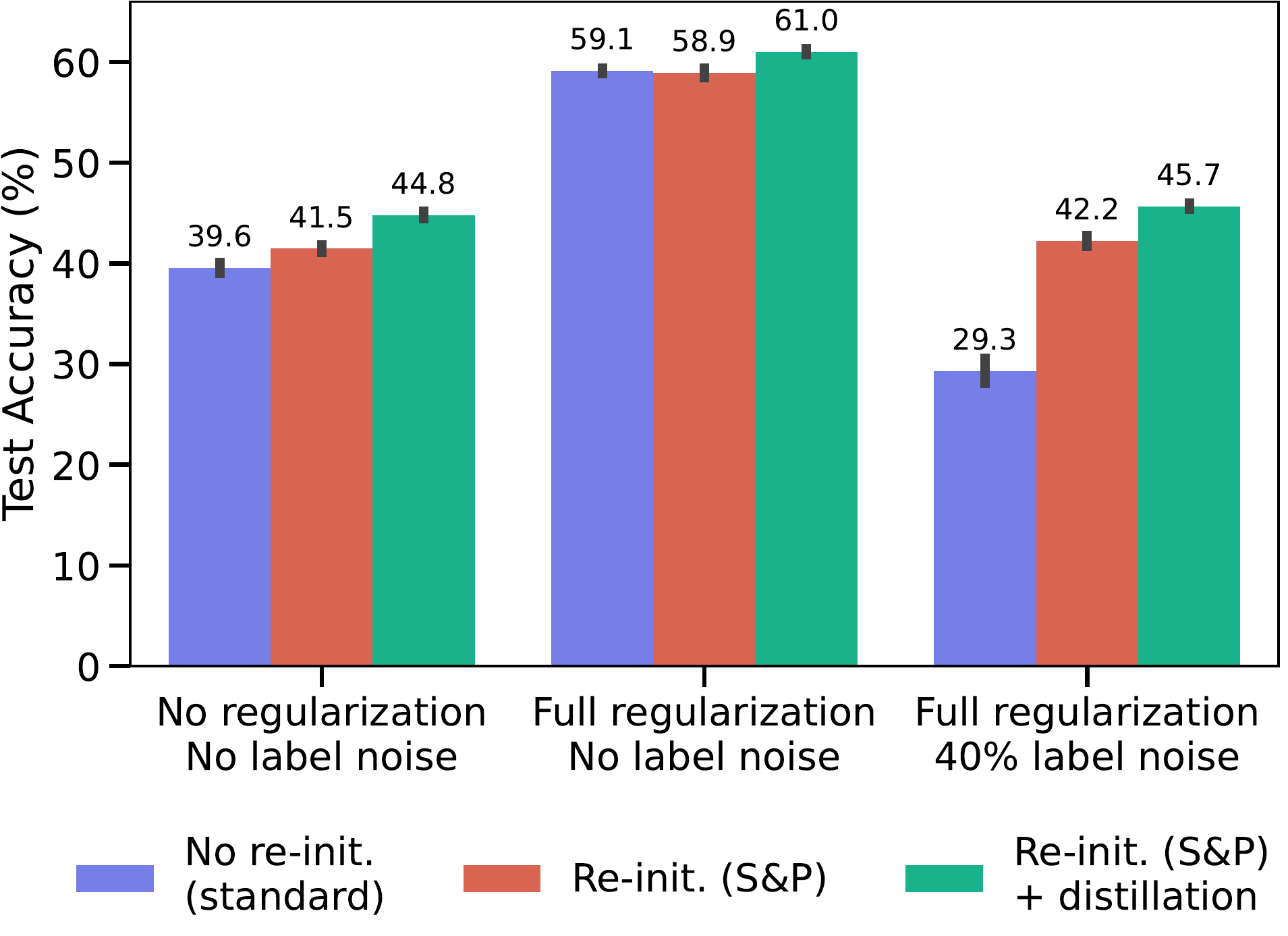}
    \vspace{-5pt}
    \caption{\textbf{Comparison of standard training (\textit{i.e.} no re-initialization) with re-initialization using \shp{} in three scenarios on \tin{} with \preresnet}. Full regularization refers to the usage of carefully tuned data augmentation, weight decay and a learning rate schedule, whereas no regularization is without. Without label noise, \shp{} helps in the absence of other regularization but has limited benefit otherwise. With label noise, \shp{} helps significantly even alongside other regularization. Adding distillation further improves performance. All training protocols have approximately equal computational cost.}\label{fig:tiny-imagenet-summary}
    \vspace{-14pt}
\end{figure}

Recent works~\citep[e.g.][]{alabdulmohsin2021impact, ash2020warmstart} have proposed a set of techniques for training neural networks based on \textit{re-initialization}. These methods, which we collectively refer to as \textit{re-initialization methods}, involve re-initializing and transforming a part or all of the parameters of a neural network periodically throughout learning. 
Studies have shown that re-initialization can help in certain settings, such as small-data regimes \citep{alabdulmohsin2021impact, taha2021knowledge} and online learning \citep{ash2020warmstart}. However, despite having no overhead in terms of computation cost and small implementation overhead, such re-initialization techniques have not yet been adopted as common deep learning practice.
Indeed, most state-of-the-art (SOTA) training protocols do not incorporate re-initialization techniques, relying instead on
advances in \textit{e.g.} optimization \citep{foret2020sharpness}, architectures \citep{liu2021swin, tan2019efficientnet, brock2021high}, data augmentation \citep{cubuk2018autoaugment, cubuk2020randaugment} and pre-training \citep{chen2020simple}.  

However, prior work suggests that re-initialization can improve the generalization performance of neural networks. For example, in the context of online learning, \citet{ash2020warmstart} studied a scenario in which training data arrives sequentially in ``chunks'' over time such that at any point in time the training dataset consists of the union of all chunks arrived so far. They compared training the network from scratch each time a new chunk arrives to \textit{warm-starting}, where we continuously train (\textit{i.e.} fine-tune) the model, finding that warm-starting significantly underperformed. In order to remedy this, they proposed \shp{} (defined in \cref{sec:sp}), a re-initialization technique applied each time a new chunk arrives. Interestingly, applying \shp{} not only closed the performance gap between warm-starting and training from scratch, but \textit{improved upon the model trained from scratch}. Does this mean that the standard approach to training neural networks without re-initialization is sub-optimal?

The motivation of our work is to systematically understand the benefits and limitations of re-initialization methods in training neural networks. To that end, we trained over \numexp{} models to identify the settings under which re-initialization methods are helpful. We study the interaction between re-initialization and other widely used regularization and optimization techniques, including data augmentation, weight decay and learning rate schedules. We seek to answer a fundamental, practical question: are the benefits of re-initialization additive with those of common, existing techniques for improving generalization? Should re-initialization be present in the arsenal of deep learning practices? Our experiments therefore include careful ablations in which we consider settings ranging from vanilla training to SOTA protocols which yield top performance for a given architecture. In addition to varying the training protocol, we also investigate the impact of re-initialization methods when learning on noisy data by studying what happens under the presence of label noise, leading to some surprising results. 

Our empirical study centers around two re-initialization methods. First, as our primary focus, we consider \textit{\shp{}}, which was proposed by \citet{ash2020warmstart} in the context of online learning and warm-starting. To the best of our knowledge, the regularization benefits of \shp{} as a re-initialization method for standard i.i.d. supervised learning have not been studied before, and we find that an adaptation of \shp{} can sometimes lead to significant gains in this setting. Second, we consider the recently proposed \textit{\lwr{}} \citep{alabdulmohsin2021impact} that has been shown to outperform prior re-initialization schemes on a variety of datasets. Furthermore, we show that re-initialization methods can naturally be adapted to incorporate self-distillation \citep{furlanello18a} which typically leads to improvement, at negligible cost. \cref{fig:tiny-imagenet-summary} summarizes our results on \tin{}. From here on, we will use \textit{standard training} to refer to training \textit{without} re-initialization.

Our contributions and findings are as follows:

\begin{itemize}[topsep=0pt,itemsep=0pt,leftmargin=*]
    \item \textbf{Shrink \& Perturb can benefit i.i.d. learning.} We investigate the benefits of \textit{\shp{}}, previously proposed for online learning, and its variant with distillation as re-initialization methods that can be used for i.i.d supervised learning. We show that they are helpful and outperform techniques such as \lwr{} in certain settings.
    \item \textbf{Re-initialization improves generalization in the absence of other regularization (\cref{sec:resets-regularization}).} There is a consistent advantage in periodically re-initializing a neural network---even up to 25 times---during training, pointing to the inherent regularization benefit of re-initialization.
    \item \textbf{Re-initialization has limited advantage over standard training in a SOTA setting (\cref{sec:reinit-with-reg}).} When data augmentation, weight decay and learning rate schedules are carefully tuned, re-initialization performs at par with standard training. However, the optimal performance becomes \textit{more robust} to the choice of learning rate and weight decay, which is desirable in practice.
    \item \textbf{Under label noise, re-initialization methods lead to significant improvement in generalization (\cref{sec:reinit-with-label-noise}).} This improvement appears on top of other well-tuned regularization, revealing a setting where the effects of re-initialization and other regularization techniques do not overlap.
   
\end{itemize}

\section{Related Work}

Various re-initialization methods have been proposed in the literature, differing usually by motivation and choice of weights that are re-initialized. For example, motivated by sparsity, \citet{han2016dsd} introduced dense-sparse-dense training, where after an initial stage of training, the smallest weights are pruned to induce sparsity, the network is trained and in the third and last stage, the pruned weights are re-initialized from zero. \citet{taha2021knowledge} also proposed ``knowledge evolution'' which partitions the weights of a network into two parts, one of which is continuously training and the other repeatedly re-initialized. For transfer learning, \citet{Li2020RIFLEBI} proposed to periodically re-initialize only the final layer, optionally with ensembling \citep{zhao2018retrain}. Recently, \citet{alabdulmohsin2021impact} compared a number of these methods with their proposed \lwr{}. They found it outperformed prior approaches, which is why we chose to study it in this work. Their empirical evaluation focused on the small-data regime, where one set of hyperparameters common across multiple datasets and architectures are used. They also did not study how re-initialization methods interact with other regularizers. Also, in the spirit of re-initialization, various approaches for ``resetting'' or ``restarting'' training can also be found in the literature \citep{loshchilov2016sgdr, huang2017snapshot, smith2017cyclical, izmailov2018swa}.

Re-initialization also recently appeared in the context of online learning and warm-starting neural networks, where data arrives sequentially \citep{ash2020warmstart, caccia2021anytime}. \citet{ash2020warmstart} showed that \textit{warm-starting} neural network training worsens generalization. They proposed \shp{}, a simple technique to improve performance in their online learning setup. The benefits of \shp{} as a generic re-initialization method for i.i.d. learning have not been studied before. \citet{caccia2021anytime} studied techniques for expanding capacity by initializing \textit{new} parameters as more data arrives. \citet{igl2021transient} observed a similar phenomenon in reinforcement learning and proposed distillation as a remedy. Generally, self-distillation \citep{furlanello18a, hinton2015distilling} itself closely relates to re-initialization (see \cref{sec:ban}).

\section{Background on Re-initialization Methods}

Let $\nn_\param$ be a neural network with parameters $\param \in \R^\paramdim$. We will assume $\nn_\param$ is trained to minimize some loss function $\loss(\param)$. Moreover, let $\initdist$ be a probability distribution over $\R^\paramdim$ from which we sample the \textit{initialization parameters} $\param_0 \sim \initdist$, where optimization begins. Then, we define a \textit{re-initialization} of $\nn_\param$ to be the function $\nn_{\reinitparam}$, where we have replaced the current parameters $\param$ with re-initialized parameters $\reinitparam = \ritransform(\newinitparam, \param)$. The function $\ritransform$ computes the re-initialized parameters using freshly initialized parameters $\newinitparam \sim \initdist$ and the current parameters $\param$. Different re-initialization methods will differ in terms of $\ritransform$.

Re-initialization methods will modify standard training as follows. Assuming a computational budget corresponding to $\numepochs$ epochs over the training data, we train the model in $\numstages$ \textit{stages}. Each stage will consist of training the network for $\lfloor \numepochs / \numstages \rfloor$ epochs. Between any two stages, we apply re-initialization as described earlier. This ensures that the computational cost in terms of the total number of gradient steps is approximately equal to the cost of standard training for $\numepochs$ epochs. This is key to ensure that the performance of different methods can be compared fairly. Explicitly, let $\param_{\stageindex-1}^\text{end}$ denote the parameters at the end of the previous stage $\stageindex-1$. Stage $\stageindex$ then consists of training the model starting from initial parameters $\param_t^\text{start} = \ritransform(\newinitparam, \param_{\stageindex-1}^\text{end})$. The initial parameters for stage 1 are sampled directly from $\initdist$. See \cref{alg:reinit} for a pseudo-code description.

\vspace{-2mm}
\subsection{Incorporating Self-Distillation With Re-initialization}\label{sec:ban}
Our experiments will also consider the effect of incorporating \textit{self-distillation}, which is a natural extension of re-initialization methods because training is split into multiple sequential stages. Specifically, when using distillation, we train the model in stage $\stageindex$ by minimizing a combination of the training loss and a distillation loss between the current model (the student) and the model at the end of the previous stage $\stageindex - 1$ (the teacher), that is, we minimize the loss
\begin{equation}
    \loss(\param) + \distillcoeff \, \infdiv{\nn_{\param_{\stageindex-1}^\text{end}}}{\nn_\param}, \label{eq:opt-obj}
\end{equation}
with respect to $\param$. The second term measures the KL-divergence between the predicted class probabilities of the teacher and the student on the training data, and $\distillcoeff$ is a hyperparameter determining the strength of distillation. In stage $\stageindex = 1$, we only optimize the loss $\loss(\param)$.
Note that this requires generating the predictions of the teacher network at the end of the previous stage and caching them for use during the optimization of the student in the current stage. The added computational cost is negligible, and we found that the training time with and without distillation did not vary much. We also remark that the setting in which re-initialization simply consists of replacing all the old model parameters with a new initialization, \textit{i.e.} $\reinitparam = \newinitparam$, combined with distillation is equivalent to Born-Again Networks (BANs) \citep{furlanello18a} supervised by the true and teacher labels, with a \textit{fixed} computational budget.\footnote{In the original BANs, the computational cost scaled proportionally with the number self-distillation iterations. For a fair comparison, we consider BANs under a fixed total budget.}

\vspace{-2mm}
\subsection{Shrink \& Perturb}\label{sec:sp}

\begin{figure}[t]
    \centering
    \includegraphics[width=0.45\linewidth]{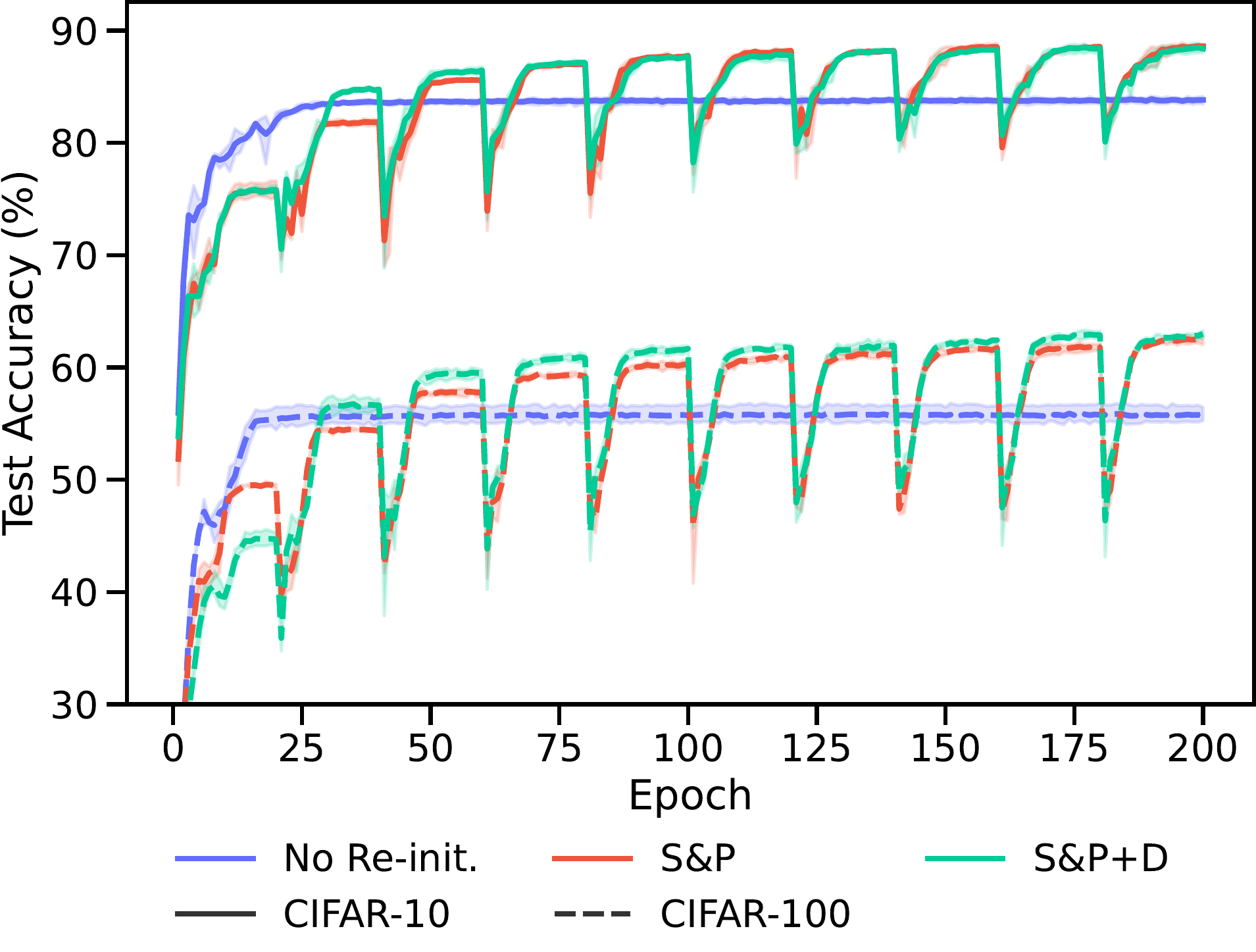}
    \vspace{-1pt}
    \caption{\textbf{Example test accuracy curves on the CIFAR datasets with ResNet-18 in setting \setA{}.} \shp{} involves 10 stages. For each method and dataset, the learning rate is tuned separately -- the optimal learning rates are on average 10 times smaller for re-initialization than standard training here.}
    \label{fig:learning-curves}
    \vspace{-5pt}
\end{figure}

Our primary focus in this work will be \shp{}, a method proposed by \citet{ash2020warmstart} in the context of online learning and warm-starting neural network training. \citet{ash2020warmstart} studied a scenario in which training data arrives sequentially in ``chunks'' over time such that at any point in time the training dataset consists of the union of all chunks arrived so far. They compared re-training the network from scratch each time a new chunk arrives to \textit{warm-starting}, where we continuously train (\textit{i.e.} fine-tune) the model. They found that the warm-started network significantly underperformed.
In order to remedy this, they proposed \shp{}, a re-initialization technique applied to the previous model each time a new chunk arrives. \shp{} consists of defining the re-initialized parameters to be: 
\begin{equation*}
    \reinitparam = \shrink \param + \perturb \newinitparam,
\end{equation*}
where $\shrink, \perturb \in [0, 1]$ are hyperparameters. We \textit{shrink} the current parameters by a multiplicative factor of $\shrink$ and \textit{perturb} them using a fresh initialization scaled by $\gamma$. In our initial experiments, we tuned these hyperparameters which then remain fixed at $\shrink = 0.4, \perturb = 0.1$ across all architectures and datasets and are similar to the optimal values in the setup of \citet{ash2020warmstart}. The number of stages $\numstages$ is a hyperparameter we vary in our experiments.

\shp{} was demonstrated to improve upon re-training from scratch in the online learning setup. \citet{ash2020warmstart} suggested that it may have a helpful regularization effect, but this effect has not been studied in any detail in the usual i.i.d. learning setting. Contrary to their setup where \shp{} is applied for online learning, we will view \shp{} as a re-initialization method for training on a stationary dataset as described earlier. 

\vspace{-2mm}
\subsection{Layer-wise Re-initialization}

\citet{alabdulmohsin2021impact} recently proposed \textit{\lwr{}}, which re-initializes the architecture block-by-block during training. Assuming that there are $\numblocks$ ``blocks'' in the architecture, we have a total of $\numstages = \numblocks$ stages during training. At the end of stage $t$, we re-initialize all layers after the $t$-th block, that is, we set:
\begin{equation*}
    \reinitparam = \param \odot \mask^{(\stageindex)} + \newinitparam \odot (1-\mask^{(\stageindex)}).
\end{equation*}

Here $\mask^{(\stageindex)} \in \{0, 1\}^\paramdim$ is a vector which masks all layers after the first $t$ blocks, \textit{i.e.} $\mask^{(\stageindex)}_i = 1$ if the $i$-th parameters belongs to the first $t$ blocks and is otherwise zero. $\odot$ denotes an element-wise product. More generally, note that we can have $\numstages = \numblocks \lwriters$ total stages, where at the end of each of the first $\lwriters$ stages $\stageindex = 1, \dots, \lwriters$, we re-initialize all layers after the first block. At the end of each of the second $\lwriters$ stages $\stageindex = \lwriters +1, \dots, 2\lwriters$, we re-initialize all layers after the second block and so forth. We note that \citet{alabdulmohsin2021impact} set $\lwriters = 1$ in their main experiments.

Moreover, in addition to re-initializing all blocks after the $t$-th block, \lwr{} also (1) re-scales the norm of the first $t$ blocks to their norm at initialization and (2) adds a normalization layer (without trainable parameters) after block $t$ whose mean and standard deviation are computed by forward passing a batch of training data. Note that unlike \shp{}, \lwr{} has a somewhat architecture-specific definition, since it requires picking \textit{a priori} what a block corresponds to in the architecture. 
In our experiments with a ResNet-18 architecture, we considered four cases: $T = K = 2$ where the network is split between the second and third residual blocks, $T = K = 5$ where we split by each residual block and lastly $T = 10$ and $20$ by letting $M = 2$ and $4$ respectively in the previous case.

\vspace{-2mm}
\section{The Regularizing Effect of Re-initialization}\label{sec:resets-regularization}

\begin{table*}[t]
\setlength\tabcolsep{2pt}
\centering
\caption{Test accuracy (\%) of different methods in settings ranging from basic to SOTA protocols on CIFAR-10 with ResNet-18.}\label{tbl:stacking-regularizers-c10}
\resizebox{\textwidth}{!}{\begin{tabular}{@{}cccccccccc@{}}
\toprule
\multirow{2}{*}{\textbf{\begin{tabular}[c]{@{}c@{}}Setting \\ Abbrev.\end{tabular}}} &
\multirow{2}{*}{\textbf{\begin{tabular}[c]{@{}c@{}}Data \\ Aug.\end{tabular}}} &
  \multirow{2}{*}{\textbf{\begin{tabular}[c]{@{}c@{}}Cosine \\ Anneal.\end{tabular}}} &
  \multirow{2}{*}{\textbf{\begin{tabular}[c]{@{}c@{}}Weight \\ Decay\end{tabular}}} &
  \multirow{2}{*}{\textbf{\begin{tabular}[c]{@{}c@{}}No Re-initialization\\(standard training)\end{tabular}}} &
  \multirow{2}{*}{\textbf{\begin{tabular}[c]{@{}c@{}}Self-distillation \\ (fixed-budget BAN)\end{tabular}}} &
  \multicolumn{2}{c}{\textbf{\begin{tabular}[c]{@{}c@{}}Layer-wise \\ Re-initialization\end{tabular}}} &
  \multicolumn{2}{c}{\textbf{Shrink \& Perturb}} \\ \cmidrule(l){7-10} 
 &
   &
   &
   &
   &
  \multicolumn{1}{c}{} &
  \multicolumn{1}{c}{w/o dist.} &
  \multicolumn{1}{c}{w/ dist.} &
  \multicolumn{1}{c}{w/o dist.} &
  \multicolumn{1}{c}{w/ dist.} \\ \midrule
\textsc{\setA} &
\xmark &
  \xmark &
  \xmark &
$83.8\,\textcolor{gray}{\pm 0.3}$ & $84.0\,\textcolor{gray}{\pm 0.4}$ & $87.6\,\textcolor{gray}{\pm 0.4}$ & $87.3\,\textcolor{gray}{\pm 0.2}$ & $\mathbf{88.8}\,\textcolor{gray}{\pm 0.2}$ & $\mathbf{88.6}\,\textcolor{gray}{\pm 0.4}$
   \\
\textsc{\setB} &
\cmark &
  \xmark &
  \xmark &
$92.5\,\textcolor{gray}{\pm 0.0}$ & $92.6\,\textcolor{gray}{\pm 0.2}$ & $92.6\,\textcolor{gray}{\pm 0.3}$ & $93.2\,\textcolor{gray}{\pm 0.1}$ & $93.1\,\textcolor{gray}{\pm 0.2}$ & $\mathbf{94.1}\,\textcolor{gray}{\pm 0.0}$
    \\
\textsc{\setC} &
\cmark &
  \cmark &
  \xmark &
$92.8\,\textcolor{gray}{\pm 0.2}$ & $92.6\,\textcolor{gray}{\pm 0.1}$ & $93.4\,\textcolor{gray}{\pm 0.1}$ & $93.4\,\textcolor{gray}{\pm 0.2}$ & $\mathbf{94.1}\,\textcolor{gray}{\pm 0.2}$ & $\mathbf{94.2}\,\textcolor{gray}{\pm 0.1}$
    \\
\textsc{\setD} &
\cmark &
  \cmark &
  \cmark &
$95.0\,\textcolor{gray}{\pm 0.0}$ & $94.8\,\textcolor{gray}{\pm 0.2}$ & $94.7\,\textcolor{gray}{\pm 0.2}$ & $95.0\,\textcolor{gray}{\pm 0.3}$ & $94.7\,\textcolor{gray}{\pm 0.2}$ & $94.7\,\textcolor{gray}{\pm 0.1}$
     \\ \bottomrule
\end{tabular}}
\end{table*}

\begin{table*}[t!]
\setlength\tabcolsep{2pt}
\centering
\caption{Test accuracy (\%) of different methods in settings ranging from basic to SOTA protocols on CIFAR-100 with ResNet-18.}\label{tbl:stacking-regularizers-c100}
\resizebox{\textwidth}{!}{\begin{tabular}{@{}ccccccccccc@{}}
\toprule
\multirow{2}{*}{\textbf{\begin{tabular}[c]{@{}c@{}}Setting \\ Abbrev.\end{tabular}}} &
\multirow{2}{*}{\textbf{\begin{tabular}[c]{@{}c@{}}Data \\ Aug.\end{tabular}}} &
  \multirow{2}{*}{\textbf{\begin{tabular}[c]{@{}c@{}}Cosine \\ Anneal.\end{tabular}}} &
  \multirow{2}{*}{\textbf{\begin{tabular}[c]{@{}c@{}}Weight \\ Decay\end{tabular}}} &
  \multirow{2}{*}{\textbf{\begin{tabular}[c]{@{}c@{}}No Re-initialization\\(standard training)\end{tabular}}} &
  \multirow{2}{*}{\textbf{\begin{tabular}[c]{@{}c@{}}Self-distillation \\ (fixed-budget BAN)\end{tabular}}} &
  \multirow{2}{*}{\textbf{\begin{tabular}[c]{@{}c@{}}SGDR\end{tabular}}} &
  \multicolumn{2}{c}{\textbf{\begin{tabular}[c]{@{}c@{}}Layer-wise \\ Re-initialization\end{tabular}}} &
  \multicolumn{2}{c}{\textbf{Shrink \& Perturb}} \\ \cmidrule(l){8-11} 
 &
   &
   &
   &
   &
  \multicolumn{1}{c}{} &
  \multicolumn{1}{c}{} &
  \multicolumn{1}{c}{w/o dist.} &
  \multicolumn{1}{c}{w/ dist.} &
  \multicolumn{1}{c}{w/o dist.} &
  \multicolumn{1}{c}{w/ dist.} \\ \midrule
\textsc{\setA{}} &
\xmark &
  \xmark &
  \xmark &
$55.5\,\textcolor{gray}{\pm 0.6}$ & $56.4\,\textcolor{gray}{\pm 0.5}$ & \textsc{n/a} & $61.0\,\textcolor{gray}{\pm 0.6}$ & $62.5\,\textcolor{gray}{\pm 0.2}$ & $63.1\,\textcolor{gray}{\pm 0.6}$ & $\mathbf{63.5}\,\textcolor{gray}{\pm 0.3}$
   \\
\textsc{\setB{}} &
\cmark &
  \xmark &
  \xmark &
$70.8\,\textcolor{gray}{\pm 0.1}$ & $70.5\,\textcolor{gray}{\pm 0.5}$ & \textsc{n/a} & $72.1\,\textcolor{gray}{\pm 0.3}$ & $\mathbf{74.7}\,\textcolor{gray}{\pm 0.2}$ & $71.9\,\textcolor{gray}{\pm 0.1}$ & $74.0\,\textcolor{gray}{\pm 0.6}$
    \\
\textsc{\setC{}} &
\cmark &
  \cmark &
  \xmark &
$71.2\,\textcolor{gray}{\pm 0.2}$ & $70.9\,\textcolor{gray}{\pm 0.4}$ & $71.0\,\textcolor{gray}{\pm 0.6}$ & $74.6\,\textcolor{gray}{\pm 0.5}$ & $\mathbf{75.4}\,\textcolor{gray}{\pm 0.2}$ & $\mathbf{75.4}\,\textcolor{gray}{\pm 0.3}$ & $\mathbf{75.4}\,\textcolor{gray}{\pm 0.4}$
    \\
\textsc{\setD{}} &
\cmark &
  \cmark &
  \cmark &
$\mathbf{77.9}\,\textcolor{gray}{\pm 0.2}$ & $77.2\,\textcolor{gray}{\pm 0.1}$ & $77.5\,\textcolor{gray}{\pm 0.2}$ & $77.5\,\textcolor{gray}{\pm 0.1}$ & $77.3\,\textcolor{gray}{\pm 0.3}$ & $77.5\,\textcolor{gray}{\pm 0.2}$ & $77.0\,\textcolor{gray}{\pm 0.3}$
     \\ \bottomrule
\end{tabular}}
\end{table*}

\begin{table*}[t]
\vspace{-3mm}
\centering
\caption{Test accuracy (\%) of standard training and \shp{} on \tin{} with \preresnet.}\label{tbl:stacking-regularizers-tin}
\resizebox{0.75\textwidth}{!}{\begin{tabular}{ccccccc}
\toprule
\multicolumn{1}{c}{\multirow{2}{*}{\textbf{\begin{tabular}[c]{@{}c@{}}Setting \\ Abbrev.\end{tabular}}}} &
\multirow{2}{*}{\textbf{\begin{tabular}[c]{@{}c@{}}Data\\ Aug.\end{tabular}}} &
\multirow{2}{*}{\textbf{\begin{tabular}[c]{@{}c@{}}Cosine\\ Anneal.\end{tabular}}} &
\multirow{2}{*}{\textbf{\begin{tabular}[c]{@{}c@{}}Weight\\ Decay\end{tabular}}} &
\multirow{2}{*}{\textbf{\begin{tabular}[c]{@{}c@{}}No Re-initialization\\(standard training)\end{tabular}}} &
\multicolumn{2}{c}{\textbf{Shrink \& Perturb}} \\ \cmidrule{6-7} 
\multicolumn{1}{c}{} &        &        &        &   & w/o dist. & w/ dist. \\ \midrule
    \textsc{\setA{}} & \xmark & \xmark & \xmark & $39.55 \textcolor{gray}{\pm 0.69}$ & $41.47 \textcolor{gray}{\pm 0.48}$ & $44.81 \textcolor{gray}{\pm 0.46}$ \\
\textsc{\setD{}} & \cmark & \cmark & \cmark & $59.12 \textcolor{gray}{\pm 0.33}$ & $58.95 \textcolor{gray}{\pm 0.56}$ & $61.03 \textcolor{gray}{\pm 0.38}$ \\
\bottomrule
\end{tabular}}
\end{table*}

In this section, we investigate the effect of re-initialization when no other regularization is deployed. We show that re-intiliazation provides a consistent and considerable advantage over standard training, even beyond the small-data regime \citep{alabdulmohsin2021impact, taha2021knowledge}, as is the case of Tiny ImageNet. This highlights their effectiveness as a regularization technique. Additionally, we find that \shp{} outperforms \lwr{} in this setting.

\vspace{-2mm}
\subsection{Experimental Setup}\label{sec:exp-setup}

For each architecture (ResNet-18, \preresnet, MobileNetV2) \citep{resnet, preact-resnet, mobilenetv2} and dataset (CIFAR-10, CIFAR-100, \tin), we used SGD with momentum $0.9$ for training with early stopping by validation accuracy. For the CIFAR datasets, we separately tuned the learning rate and weight decay hyperparameters for each method by cross validation. Note that this is different to the setup of \citet{alabdulmohsin2021impact} where the learning rate and weight decay hyperparameters were fixed for all combinations of datasets and architectures in their image classification experiments.
For the larger and more computationally costly \tin{} dataset, we tuned the learning rate and weight decay hyperparameters for standard training (that is, no re-initialization) and kept them fixed for the re-initialization methods. This simulates a practical scenario: given tuned hyperparameters for standard training on an expensive and large dataset, does it help to use re-initialization without re-tuning the hyperparameters?

For re-initialization methods, we note that the number of stages $T$ is an additional hyperparameter which we tune (\textit{cf.} \cref{sec:num-stages}). Moreover, all models are optimized for the same number of gradient steps for all methods. All other hyperparameters such as batch size and number of epochs are the same for all methods and specified in \cref{tbl:hyperparams-cifar,tbl:hyperparams-tiny} of \cref{sec:hyperparameters}. Our code will be open-sourced.

\subsection{Re-initialization Improves Performance in the Absence of Other Regularization}
The different settings we consider are labeled \setA{}, \setB{}, \setC{} and \setD{}, as described on the left of \cref{tbl:stacking-regularizers-c10}. Setting \setA{} is the basic setting where we use a constant learning rate, no weight decay and no data augmentation. As shown in the first rows of \cref{tbl:stacking-regularizers-c10,tbl:stacking-regularizers-c100}, both \shp{} and \lwr{} improve generalization performance over standard training for both CIFAR datasets. Moreover, \shp{} outperforms \lwr{} by a margin of 1-2 percentage points in both cases. Both re-initialization methods also benefit from incorporating distillation for CIFAR-10, although distillation matters less for CIFAR-100. Next, we consider training in setting \setA{} for \tin{}, focusing only on \shp{} with and without distillation in this case, as shown in \cref{tbl:stacking-regularizers-tin}. As before, re-initialization is beneficial compared to standard training, particularly when we include distillation. Indeed, these results indicate that re-initialization yields a notable generalization boost in setting \setA{} without additional computational cost.

Examples of the test accuracy learning curves are shown in \cref{fig:learning-curves} for \shp{} with 10 stages. Notice that each re-initialization causes a sudden drop in performance from which there is a quick recovery. Whereas the test accuracy for standard training stagnates early on, training with \shp{} keeps improving test accuracy with each successive re-initialization and already outperforms standard training by epoch 50. All models shown in this plot achieved approximately 100\% training accuracy and close to zero training negative log-likelihood. See \cref{fig:train-acc-curves,fig:train-loss-curves} for examples of training curves.

\vspace{-5pt}
\section{Re-initialization Alongside Other Regularization}\label{sec:reinit-with-reg}

\cref{sec:resets-regularization} demonstrates that \reinit{} methods have a regularizing effect on learning that improves generalization. Next, we consider how beneficial \reinit{} is when deployed alongside other regularization techniques that are commonly used to achieve high performance in SOTA training protocols. In particular, we consider three common regularization techniques: data augmentation, cosine annealing and weight decay. Note that while cosine annealing is not an explicit regularizer, the choice of learning rates affects regularization \citep{li2019towards, li2019exponential} and is an important ingredient for well-performing models. For re-initialization methods, we apply cosine annealing \textit{per stage}, yielding a cyclical learning rate \citep{loshchilov2016sgdr}. Our motivation is to explore whether the effect of these different techniques is \textit{additive} with the benefit of \reinit{}; indeed, regularizers that are helpful each on their own are not necessarily helpful in the presence of one another. 

\vspace{-5pt}
\subsection{Re-initialization Offers Little Benefit in SOTA Settings}
Using a similar setup as before, we compare standard training with various \reinit{} methods across these settings on the CIFAR datasets in \cref{tbl:stacking-regularizers-c10,tbl:stacking-regularizers-c100}. Beginning with the simple setting explored in \cref{sec:resets-regularization}, we gradually add \textbf{d}ata augmentation (setting \setB{}), \textbf{c}osine annealing (setting \setC{}) and \textbf{w}eight decay (setting \setD{}), achieving around SOTA performance in setting \setD{}. First, we observe that all \reinit{} methods except fixed-budget BAN noticeably improve upon standard training in settings \setA{}, \setB{} and \setC{}.
In most cases, the best performing method is \shp{} with distillation, improving accuracy over standard training by upto 5 percentage points on CIFAR-10 and 8 percentage points on CIFAR-100.

\begin{figure*}[t!]
    \centering
    \includegraphics[width=0.97\linewidth]{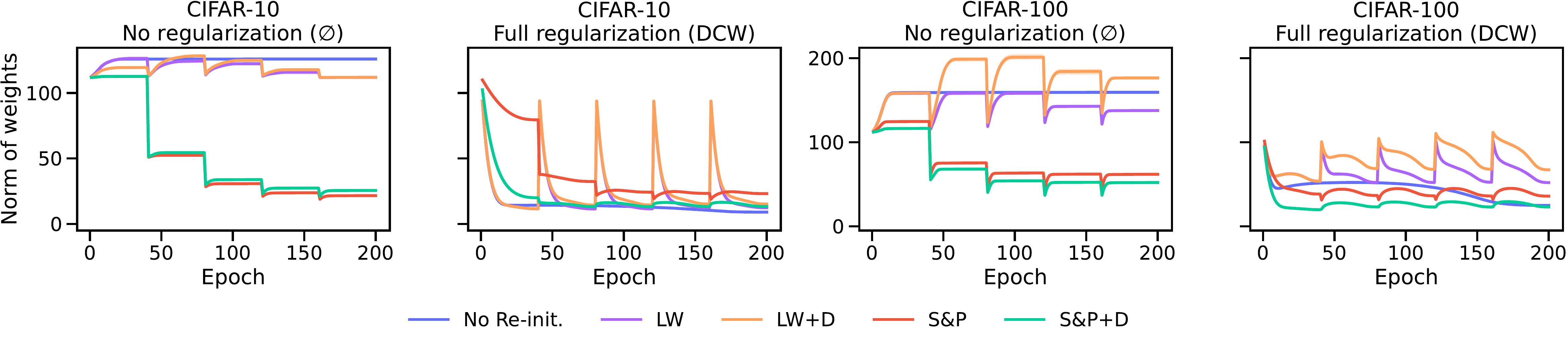}
    \caption{\textbf{Weight norms during training vary across methods more in setting \setA{} than setting \setD{}.} Curves show how the norm of ResNet-18 weights varies through training for re-initialization methods (with 5 stages) and standard training. In setting \setA{}, \shp{} yields a model with smaller weight norm than standard training and \lwr{}, but in setting \setD{}, all methods yield models with similar weight norms.}\label{fig:pnorms}
    \vspace{-5pt}
\end{figure*}

However, a key finding here is that \textit{when all regularization techniques are used and carefully tuned (setting \setD{}), the best \reinit{} methods do not improve generalization performance over standard training}. Therefore, stacking \reinit{} on top of other regularization techniques results in a \textit{sub-additive} effect on performance. Note that this setting is arguably more important and common in practice than previous settings.
We speculate that \shp{} affects the norm of the weights during training which may overlap with the effect of weight decay. As shown in \cref{fig:pnorms}, in the absence of weight decay, the norm of the weights tends to increase monotonically without re-initialization, whereas \shp{} periodically reduces this weight norm. In setting \setA{}, this results in the learned models' weight norms varying across methods, while, in setting \setD{}, all methods learn models with similar final norms due to weight decay. Moreover, in setting \setA{}, \shp{} results in a model with norm similar to that of standard training in setting \setD{}. Therefore, both weight decay and \shp{} encourage the final learned model to have a small weight norm, a quantity that is known to be linked to generalization performance \citep{neyshabur2018a, bartlett2017spectral}. Nevertheless, this does not fully explain why \shp{} works, as shown later in \cref{sec:reinit-with-label-noise}.
We also show the impact of re-initialization in setting \setD{} for \tin{} in the last row of \cref{tbl:stacking-regularizers-tin}. Similar to before, the improvement in performance from using \shp{} is much smaller here than in setting \setA{}. \shp{} has no impact on performance, whereas \shp{} with distillation slightly improves performance over standard training.

\vspace{-5pt}
\subsection{The Role of Self-Distillation}

Another observation from \cref{tbl:stacking-regularizers-c10,tbl:stacking-regularizers-c100} is that fixed-budget BANs do not improve performance compared to standard training in most cases, sometimes leading to worse test accuracies, and are outperformed by \shp{} and \lwr{} in all settings except \setD{}. Recall from \cref{sec:ban} that BANs can be viewed as a simple re-initialization method combined with distillation: we re-initialize the full network \textit{from scratch} in each stage. Therefore, BANs constitute an important baseline for more sophisticated re-initialization methods. Our results indicate that while distillation can boost performance, it is computationally sub-optimal to \textit{completely} re-initialize each network in the sequence. Under a fixed-budget setting as shown here, too many stages will then lead to too few gradient steps per stage, whereas too few stages do not reap the full benefits of distillation, both of which can negatively impact performance. On the other hand, \lwr{} and \shp{} partly ``re-use'' the model from the previous stage, circumventing the need for a large number of gradient steps per stage and improving learning efficiency. Therefore, re-initialization is itself crucial, and distillation does not suffice on its own. \cref{fig:learning-curves-bans} shows how complete re-initialization in BANs leads to large drops in performance from which it is more difficult to recover quickly.

\vspace{-5pt}
\subsection{The Role of Cosine Annealing}

Recall that cosine annealing is applied per stage for re-initialization methods, yielding a cyclical learning rate schedule as used in SGDR \citep{loshchilov2016sgdr}. Such a learning rate schedule can have its own benefits, therefore we compared re-initialization in settings \setC{} and \setD{} with SGDR for CIFAR-100. As shown in \cref{tbl:stacking-regularizers-c100}, SGDR does not match the performance of the re-initialization methods in setting \setC{} and performs similar in setting \setD{}. This indicates that when re-initialization is helpful, it is not an effect of the cyclical schedule.

\vspace{-5pt}
\subsection{Re-initialization Makes Optimal Performance More Robust to Hyperparameters}

\begin{figure*}[t!]%
     \centering%
    \includegraphics[width=0.98\textwidth]{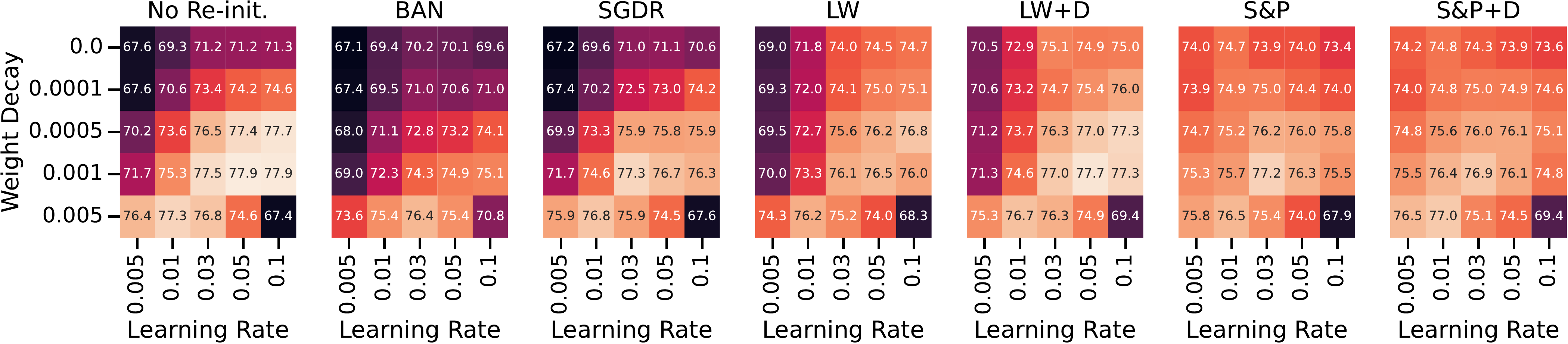}
    \caption{\textbf{Re-initialization can make performance less sensitive to the choice of learning rate and weight decay.} Performance of different methods over a grid search of learning rate and weight decay combinations on CIFAR-100 with ResNet-18 in setting \setD{}. All re-initialization methods use 5 stages. \shp{} stands out most, as its performance varies much less over the grid than other methods, especially standard training (no re-initialization).}\label{fig:grid-search-lr-wd}
    \vspace{0pt}
\end{figure*}

\begin{figure*}[t]
    \centering
    \includegraphics[width=0.95\linewidth]{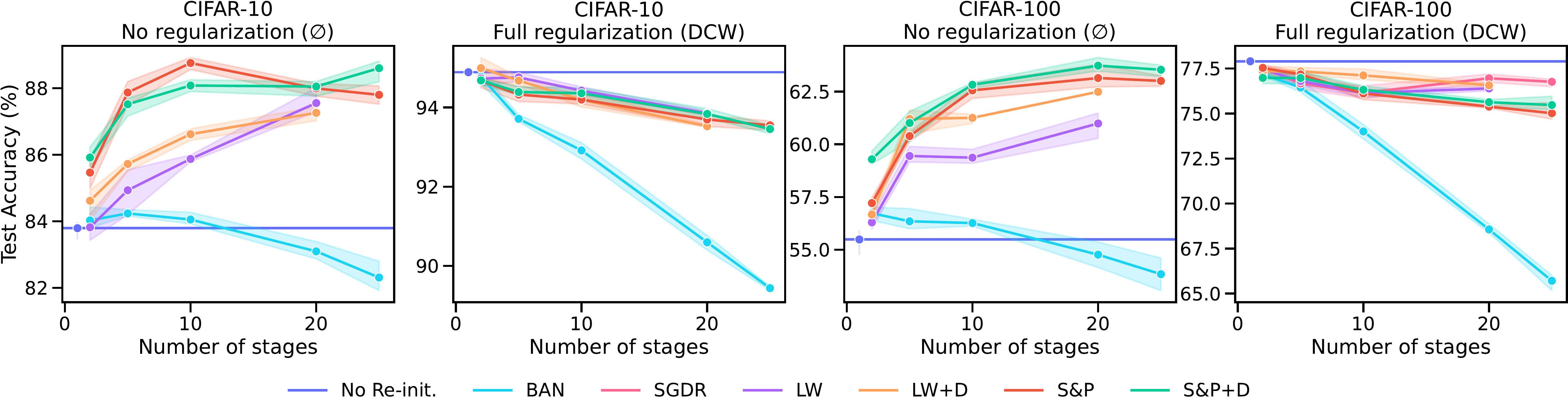}
    \caption{\textbf{Test accuracy as a function of the number of re-initialization stages in settings \setA{} and \setD{}.} In setting \setA{}, \textit{any} number of stages improves upon standard training for \shp{} and \lwr{}, though the optimal number tends to be 5-20 stages. In setting \setD{}, adding more stages monotonically worsens generalization. Recall that for \lwr{}, the number of stages is tied to the architecture itself, and $\numstages \in \{2, 5, 10, 20\}$.}\label{fig:stages}
    \vspace{-1pt}
\end{figure*}

Although re-initialization methods do not offer much benefit in terms of generalization for optimal hyperparameters in setting \setD{}, we found that they can make performance less sensitive to the choice of learning rate and weight decay hyperparameters -- particularly \shp. \cref{fig:grid-search-lr-wd} shows the test accuracy achieved by different methods over a grid of learning rate and weight decay choices for CIFAR-100 in setting \setD{}. We observe that the performance of both \lwr{} and \shp{} varies less over the grid compared to no re-initialization. This is especially prominent for \shp{}. For example, with learning rate 0.005 and weight decay 0, performance of standard training degrades to 67.6\%, whereas the performance of \shp{} remains above 73\% throughout.
Therefore, re-initialization methods can be beneficial in regimes where thoroughly hyperparameter tuning is infeasible and robustness becomes crucial. This complies with the findings of \citet{alabdulmohsin2021impact} who operated under the setup of using a common set of hyperparameters for all datasets and architectures, which may be sub-optimal in certain cases for standard training, and found re-initialization methods outperform standard training.

\vspace{-5pt}
\subsection{Impact of the Number of Re-initialization Stages}\label{sec:num-stages}

Recall that the number of stages $\numstages$ used for re-initialization is an additional hyperparameter that we tuned in our experiments. Note that $\numstages = 1$ trivially corresponds to standard training (\textit{i.e.} no re-initialization). We show how test accuracy varies as a function of $\numstages$ in \cref{fig:stages} for the different methods over each of the CIFAR datasets in settings \setA{} and \setD{}. In setting \setA{}, typically a large number of stages, in the range 5-20, is most beneficial for the re-initialization methods. In fact, for \shp{} and \lwr{}, \textit{any} number of stages in the range shown (2-25) improves upon standard training. However, in setting \setD{}, the profile of the curves changes drastically. Increasing the number of stages monotonically lowers the test accuracy, which is consistent with our finding that in this setting, re-initialization methods offer little advantage. See \cref{fig:mobilenet-stages} in \cref{sec:additional-results} for further experiments with MobileNetV2 on the CIFAR datasets.

\vspace{-3mm}
\section{Re-initialization Under Label Noise}\label{sec:reinit-with-label-noise}

\begin{figure}[t!]
    \centering
    \includegraphics[width=1.\linewidth]{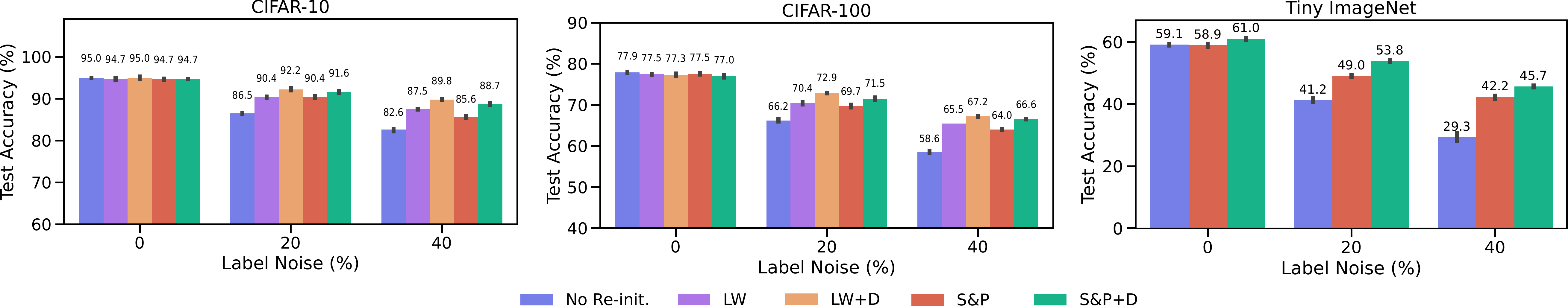}
    \vspace{-8pt}
    \caption{\textbf{Re-initialization is beneficial for learning under label noise.} Both re-initialization methods improve performance, with distillation leading to a further improvement. Moreover, the benefit of re-initialization compared to standard training increases with more label noise.}\label{fig:label-noise}
    \vspace{-8pt}
\end{figure}

Having studied how different components of the training protocol interact with re-initialization, we next focus on learning under noisy data. In particular, we add label noise to the data which makes learning more challenging. We consider the impact on performance when a randomly-chosen fraction $\lnfrac$ of the training data points have random labels, whereas the test set remains the same. The addition of noise makes learning more difficult as models would require more regularization to prevent them from overfitting to the noise. Therefore, learning under label noise can help distinguish different regularization techniques. We remain in setting \setD{}, where data augmentation, weight decay and learning rate schedules are used. Moreover, we follow the setup described in \cref{sec:exp-setup}; on the CIFAR datasets, learning rate and weight decay are tuned separately for each method, while on \tin{}, we use the hyperparameters obtained from tuning on standard training only.

\cref{fig:label-noise} shows the performance of different methods as we add $\lnfrac = 20\%$ and $40\%$ label noise on the CIFAR and \tin{} datasets. Surprisingly, re-initialization methods consistently improve upon standard training here. For CIFAR datasets, both \shp{} and \lwr{} are beneficial each on their own, and performance further improves with the addition of distillation. The results are similar for \tin{}, where the difference between standard training and re-initialization is even starker: \shp{} with distillation improves upon standard training by over 15 percentage points in test accuracy when we have 40\% label noise.

We can therefore conclude that the effects of re-initialization do not completely overlap with those of data augmentation, weight decay and learning rate schedules. Indeed, even though re-initialization does not improve generalization in setting \setD{} without label noise, it shows a substantial improvement on generalization with label noise on top of standard regularization techniques. We suspect that one reason re-initialization helps in this scenario relates to the order in which neural networks learn training examples. ``Difficult examples'' (such as mislabelled inputs) tend to be learned \textit{later} during training by \textit{deeper} layers \citep{baldock2021difficulty}. Re-initialization might be exploiting this property of learning by naturally selecting the clean data: since periodically re-initializing ``resets'' training to an extent, the network might be prevented from confidently fitting to the noise (which is learned later) to instead focus on learning the correctly labelled examples (which are learned earlier). This may also explain why \lwr{} outperforms \shp{} here due to its layer-wise structure that re-initializes deeper layers more. This also raises the question of whether the performance gap between re-initialization methods and standard training can be closed by tuning the epoch budget to reduce overfitting. As shown in \cref{fig:epoch-tuning}, this is \textit{not} sufficient to close to gap.

\vspace{-5pt}
\section{Conclusion, Limitations \& Future Work}
We have investigated when re-initialization methods, a set of techniques that are simple to implement with almost no computational overhead, improve generalization compared to standard training. We found that such methods are beneficial for improving performance in the absence of other regularization. However, in a setting resembling SOTA training protocols (\textit{i.e.} including data augmentation, weight decay and learning rate schedules), which is arguably more important and common, re-initialization methods do not improve over standard training, although optimal performance becomes more robust to the choice of hyperparameters. Finally, under label noise, re-initialization does have a significant and helpful effect on learning that other regularization techniques are not able to offer.

In future work, we hope to study the implications of our work on online learning, where \shp{} was first proposed and shown to be helpful. Does the generalization gap due to warm-starting still appear if models are trained with sufficient regularization? Does \shp{} continue to be beneficial?
One limitation of our work is that, although we observe clear empirical trends in \textit{when} re-initialization works, a deeper understanding of \textit{why} it does or does not work is missing and would be very desirable to gain in future work.  
Another limitation of our study is that we restrict ourselves to CIFAR-10/100 and \tin{} datasets and convolution-based network architectures for classification. Although this restriction allowed us to thoroughly explore the interaction between re-initialization and other techniques on these standard benchmarks and carefully tune hyperparameters given a limited compute budget, it would be interesting to extend the scope of our study to other architectures, tasks and data modalities as future work.

\bibliography{references}
\bibliographystyle{icml2022}

\newpage
\appendix
\onecolumn

\section{Additional Results}\label{sec:additional-results}

\subsection{Results on ImageNet}\label{sec:imagenet}

Our findings on the dependence of the benefits of \shp{} on other regularization were shown to hold for the CIFAR datasets and \tin{}. In this section, we preliminarily explore whether they also hold for ImageNet. As \cref{tbl:imagenet} shows, in setting \textsc{CW}, which lacks data augmentation, \shp{} clearly outperforms no re-initialization. However, in setting \setD{} with full regularization, \shp{} even slightly underperforms compared to standard training. These conclusions also match our findings in \cref{sec:resets-regularization,sec:reinit-with-reg}.

\begin{table*}[t]
\vspace{-3mm}
\centering
\caption{Test accuracy (\%) of standard training and \shp{} on ImageNet with ResNet-18.}\label{tbl:imagenet}
\resizebox{0.75\textwidth}{!}{\begin{tabular}{cccccc}
\toprule
\multicolumn{1}{c}{\multirow{2}{*}{\textbf{\begin{tabular}[c]{@{}c@{}}Setting \\ Abbrev.\end{tabular}}}} &
\multirow{2}{*}{\textbf{\begin{tabular}[c]{@{}c@{}}Data\\ Aug.\end{tabular}}} &
\multirow{2}{*}{\textbf{\begin{tabular}[c]{@{}c@{}}Cosine\\ Anneal.\end{tabular}}} &
\multirow{2}{*}{\textbf{\begin{tabular}[c]{@{}c@{}}Weight\\ Decay\end{tabular}}} &
\multirow{2}{*}{\textbf{\begin{tabular}[c]{@{}c@{}}No Re-initialization\\(standard training)\end{tabular}}} &
\multirow{2}{*}{\textbf{\begin{tabular}[c]{@{}c@{}}Shrink \& Perturb\end{tabular}}} \\
\multicolumn{1}{c}{} &        &        &        &      & \\ \midrule
    \textsc{CW} & \xmark & \cmark & \cmark & $59.1$ & $63.7$ \\
\textsc{DCW} & \cmark & \cmark & \cmark & $70.8$ & $69.7$ \\
\bottomrule
\end{tabular}}
\end{table*}

\subsection{Implications for Online Learning}\label{sec:online-learning}

\begin{figure*}[h!]
    \centering
    \includegraphics[width=0.6\linewidth]{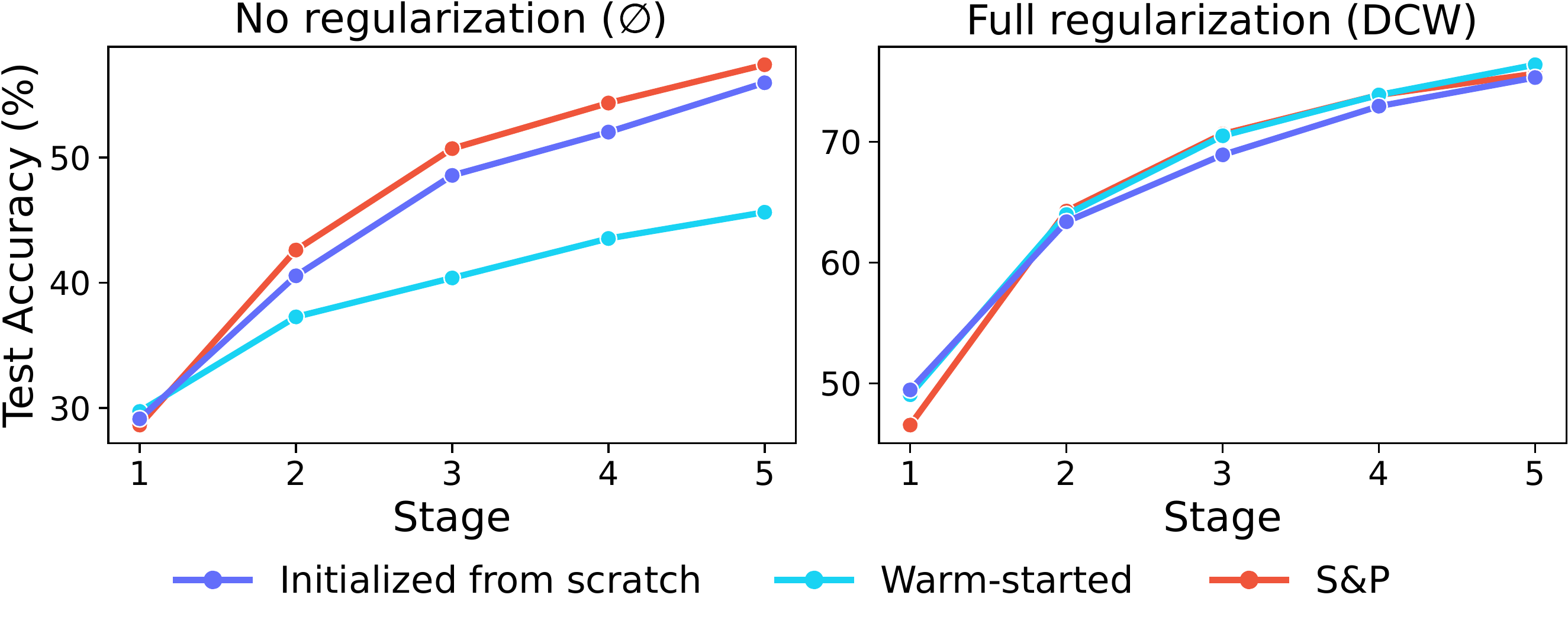}
    \vspace{-5pt}
    \caption{\textbf{The benefits of \shp{} and the warm-starting generalization gap in the online learning setup of \citet{ash2020warmstart} also depend on whether other regularization is used.} Results shown are for a ResNet-18 trained on CIFAR-100. See \cref{sec:online-learning} for discussion.}
    \label{fig:online-learning}
    \vspace{-5pt}
\end{figure*}

\shp{} was originally proposed by \citet{ash2020warmstart} in the context of online learning (\textit{cf.} \cref{sec:sp}), so we explore whether our conclusions also hold there. In particular, how do different methods compare in the absence of other regularization (similar to the setting originally studied by \citet{ash2020warmstart}) and with full regularization? In \cref{fig:online-learning}, we assume that training data accumulates by arriving sequentially in five ``chunks'' leading to five stages of training, and we compare three methods: (1) initializing the network from scratch and re-training it every time a new chunk arrives, (2) warm-starting it (\textit{i.e.} continuously fine-tuning) and (3) applying \shp{} every time a new chunk arrives. In line with the findings of \citet{ash2020warmstart}, without regularization, the warm-started network significantly underperforms the network initialized from scratch, whereas applying \shp{} outperforms both (left plot in \cref{fig:online-learning}). However, under sufficient regularization, all three methods interestingly perform approximately the same (right plot in \cref{fig:online-learning}). This also closely matches our findings from \cref{sec:resets-regularization,sec:reinit-with-reg}. 

\begin{figure*}[h!]
    \centering
    \includegraphics[width=0.42\linewidth]{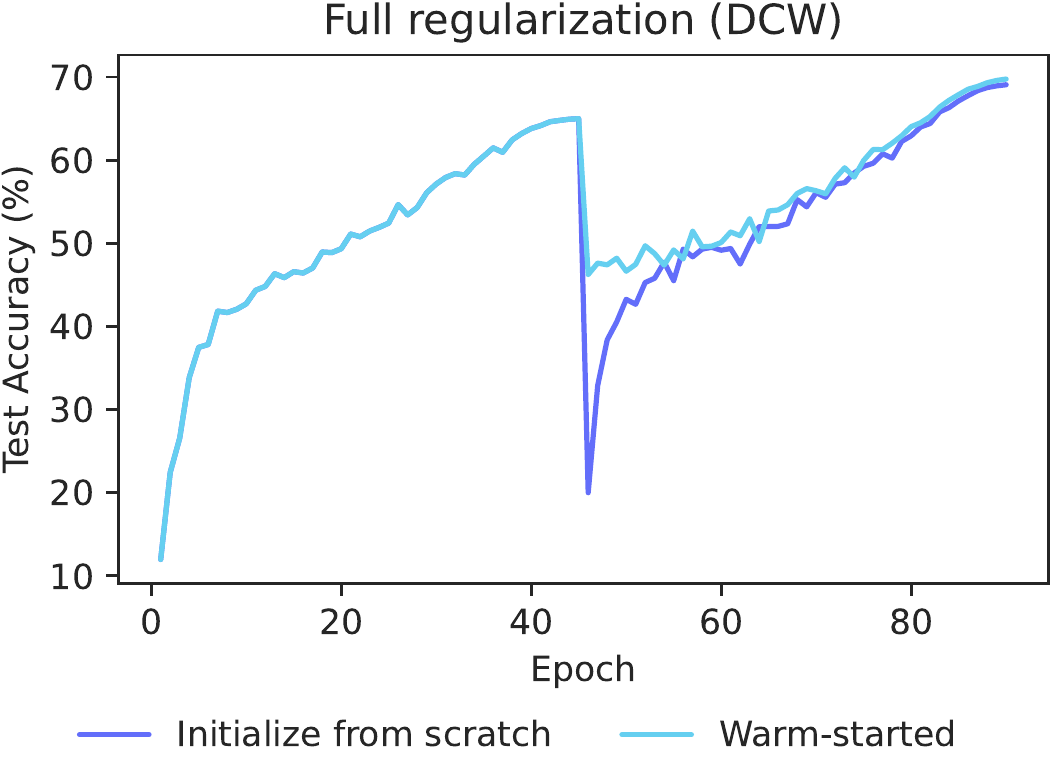}
    \vspace{-5pt}
    \caption{\textbf{There is no warm-starting generalization gap for ResNet-18 on ImageNet in the online learning setup of \citet{ash2020warmstart} under full regularization.} Learning curves show the test accuracy on ImageNet where in the first stage we train on half the training data, followed by the full training data in the second stage. See \cref{sec:online-learning} for discussion.}
    \label{fig:online-learning-imagenet}
    \vspace{-5pt}
\end{figure*}

We also find that the warm-starting generalization gap \citep{ash2020warmstart} (\textit{i.e.} warm-started models underperforming models initialized from scratch in each stage) does not exist on ImageNet in a setting with full regularization as shown in \cref{fig:online-learning-imagenet}.

\subsection{Additional Figures}

\begin{figure}[h!]
    \centering
    \includegraphics[width=0.4\linewidth]{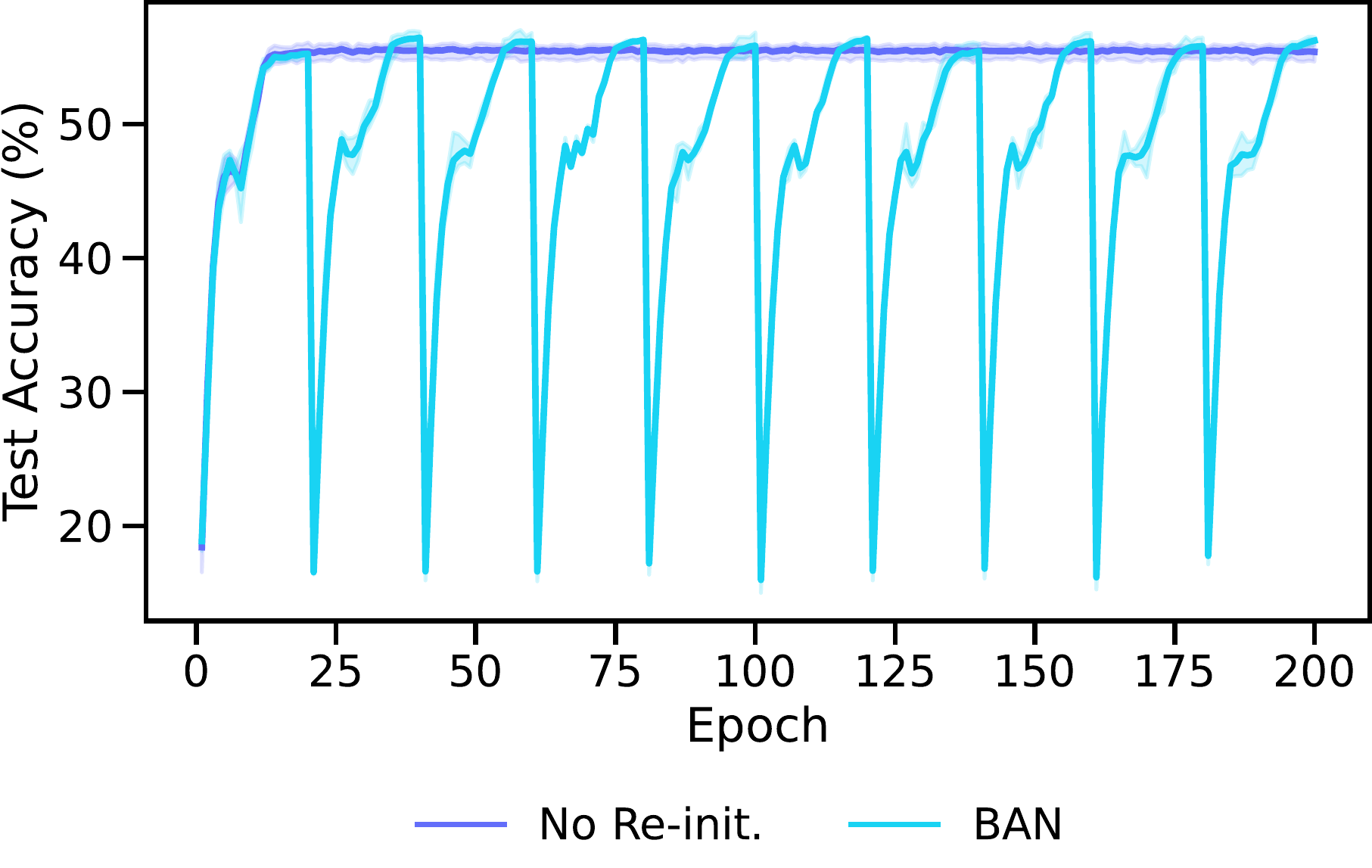}
    \vspace{-5pt}
    \caption{\textbf{Example test accuracy curves on the CIFAR-100 dataset with ResNet-18 in setting \setA.} BAN involves 10 stages. For each method and dataset, the learning rate is tuned separately. Notice that each re-initialization causes a large drop in performance in contrast with \textit{e.g.} \shp{} as shown in \cref{fig:learning-curves}.}
    \label{fig:learning-curves-bans}
    \vspace{-5pt}
\end{figure}

\begin{figure*}[h!]
    \centering
    \includegraphics[width=0.95\linewidth]{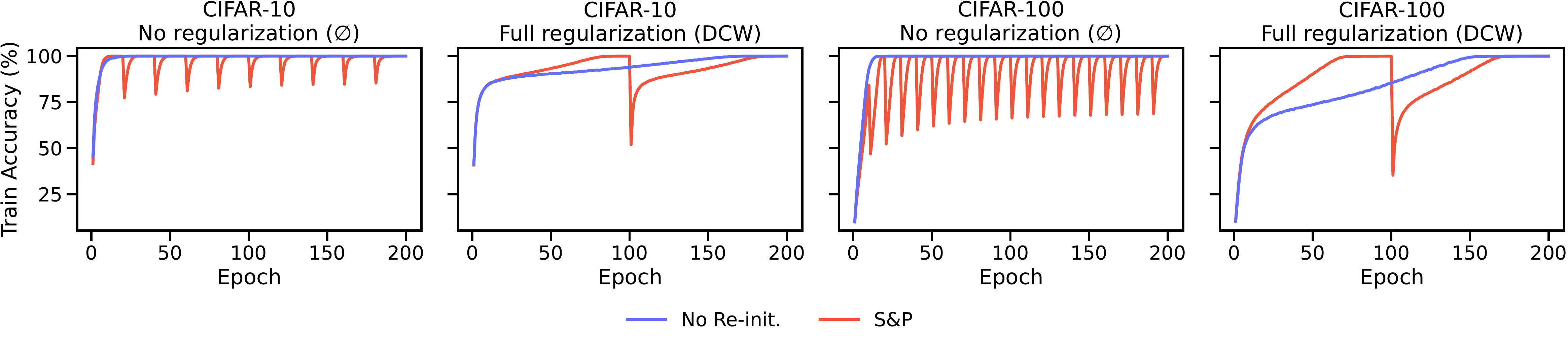}
    \vspace{-5pt}
    \caption{\textbf{Example train accuracy curves for standard training and \shp{}.} In each plot, \shp{} is shown with its optimal number of stages. Observe that in all cases the models achieve close to 100\% accuracy.}
    \label{fig:train-acc-curves}
    \vspace{-5pt}
\end{figure*}

\begin{figure*}[h!]
    \centering
    \includegraphics[width=0.95\linewidth]{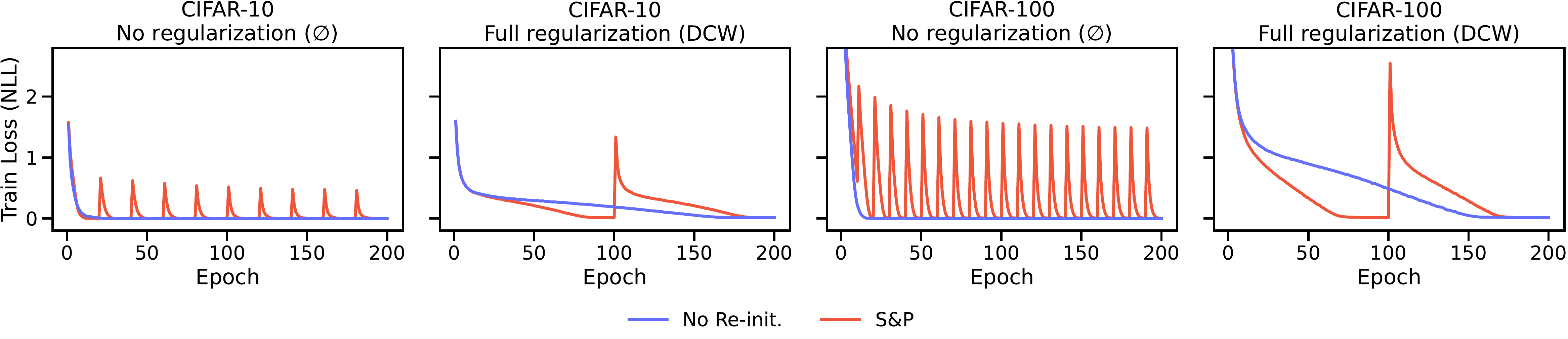}
    \vspace{-5pt}
    \caption{\textbf{Example train negative log-likelihood curves for standard training and \shp{}.} In each plot, \shp{} is shown with its optimal number of stages. Observe that in all cases the models achieve close to zero loss.}
    \label{fig:train-loss-curves}
    \vspace{-5pt}
\end{figure*}

\begin{figure*}[h!]
    \centering
    \includegraphics[width=0.95\linewidth]{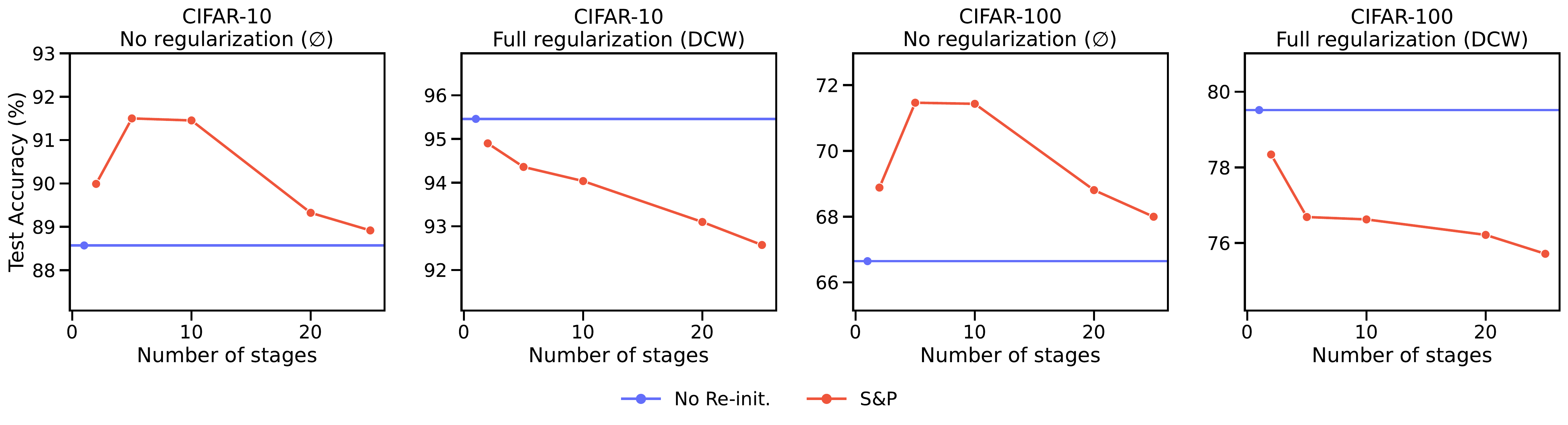}
    \vspace{-5pt}
    \caption{\textbf{Effect of \shp{} on the CIFAR datasets in settings \setA{} and \setD{} with a MobileNetV2 architecture.} Our results in this figure indicate that the overall trend observed in our experiments with ResNet-18 also appears for MobileNetV2. In particular, in setting \setA, \shp{} improves performance for any number of stages shown, whereas in setting \setD{}, it provides no benefit over standard training (no re-initialization).}
    \label{fig:mobilenet-stages}
    \vspace{-5pt}
\end{figure*}

\begin{figure*}[h!]
    \centering
    \includegraphics[width=0.9\linewidth]{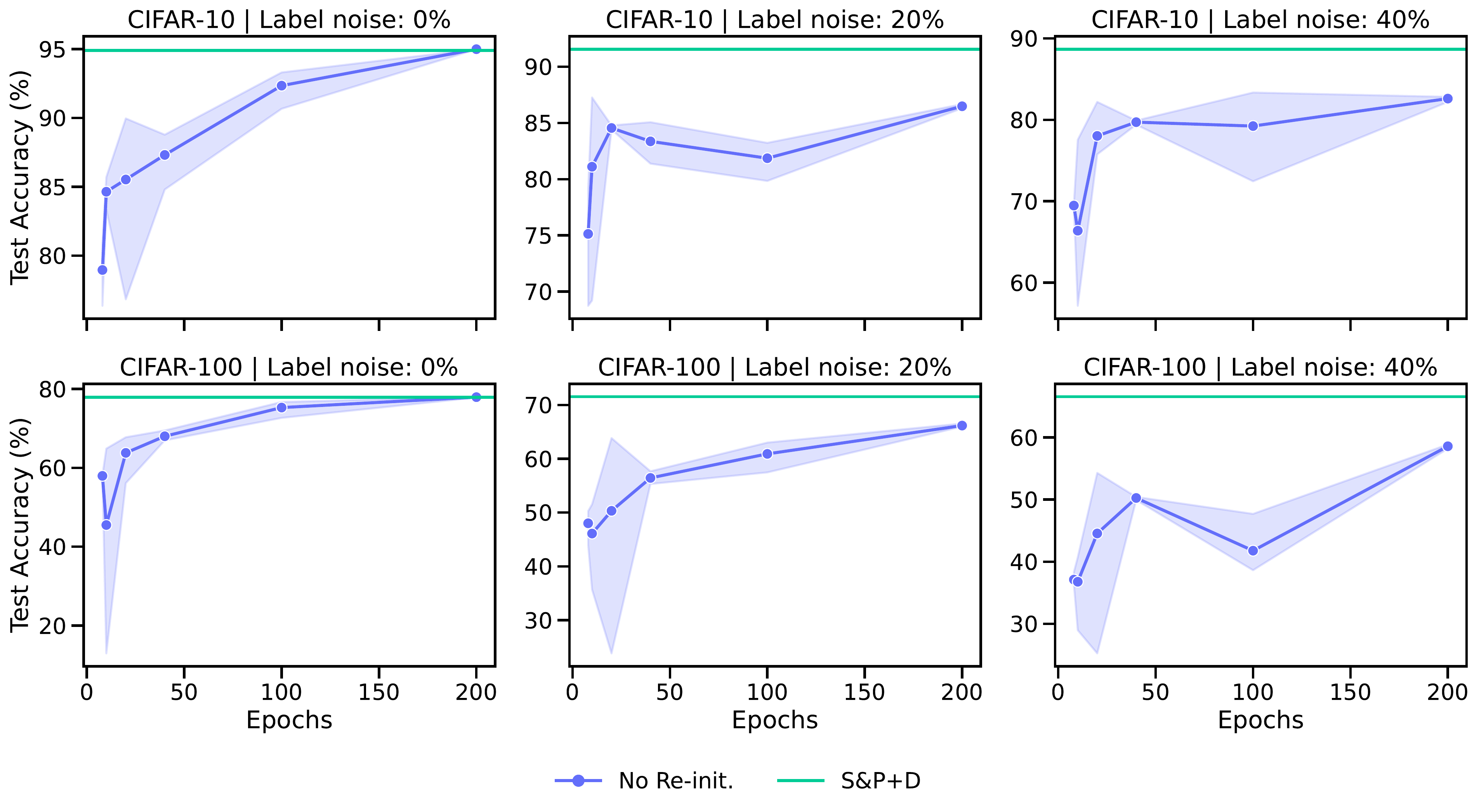}
    \vspace{-5pt}
    \caption{\textbf{Under label noise, tuning the epoch budget for standard training does not close the gap with re-initialization methods.} See the discussion in  \cref{sec:reinit-with-label-noise}.}
    \label{fig:epoch-tuning}
    \vspace{-5pt}
\end{figure*}

\cref{fig:epoch-tuning,fig:learning-curves-bans,fig:train-acc-curves,fig:train-loss-curves,fig:mobilenet-stages} are discussed in \cref{sec:reinit-with-reg,sec:resets-regularization,sec:reinit-with-label-noise}.

\begin{algorithm}[hb]
   \caption{Training with re-initialization function $\ritransform$.}\label{alg:reinit}
\begin{algorithmic}
   \STATE {\bfseries Input:} Training data $\mathcal{D}$, number of re-initialization stages $\numstages$, initialization distribution $\initdist$, total epochs $\numepochs$.
   \STATE $\epochsperstage$ $\gets \lfloor \numepochs / \numstages \rfloor$ \LineComment{\# epochs per stage}
   \FOR{$\stageindex=1$ {\bfseries to} $\numstages$}
   \IF{$t == 1$} 
   \STATE $\param^\text{start} \sim \initdist$ \LineComment{\# initialize training}
   \ELSE
   \STATE $\param^\text{start} \gets  \ritransform(\newinitparam, \param^\text{end})$, where $\param_\text{init} \sim \initdist$ is an i.i.d. initialization.  \LineComment{\# re-initialize network}
   \ENDIF
    \STATE $\param^\text{end} \gets$ parameters after training for $\epochsperstage$ epochs on $\mathcal{D}$, starting from $\param^\text{start}$, optionally with distillation from teacher network $\nn_{\param^\text{end}}$ if $\stageindex > 1$. \LineComment{\# optimization objective is defined in \cref{eq:opt-obj}}   
   \ENDFOR
    \STATE {\bfseries return} $\param^\text{end}$
\end{algorithmic}
\end{algorithm}

\section{Hyperparameters \& Pseudo-code of Re-initialization Methods}
\label{sec:hyperparameters}
Hyperparameters for our experiments are described in \cref{tbl:hyperparams-cifar,tbl:hyperparams-tiny}, and the pseudo-code for training with re-initialization is shown in \cref{alg:reinit}.

\begin{table*}[h]
    \centering
    \begin{tabular}{l c}
    \toprule
    \multicolumn{2}{c}{\textbf{General}}\\
    \midrule
    Total training epochs & $200$ \\
    Stages $\times$ epochs per stage & $\left\lbrace 2 \times 100, 5 \times 40, 10 \times 20, 20 \times 10, 25 \times 8\right\rbrace$ \\
    Batch size     &  $125$ \\
    Momentum     & $0.9$ \\
    Learning rate grid & $\left\lbrace 0.005, 0.01, 0.03, 0.05, 0.1 \right\rbrace$ \\
    Weight decay grid & $\left\lbrace 0, 0.0001, 0.0005, 0.001, 0.005 \right \rbrace$ \\
    Distillation strength $\distillcoeff$ (if used) & $1$ \\
    \multirow{2}{*}{Data augmentation (if used)} & Random horizontal flips and\\ & size-preserving crops after padding 4 pixels. \\
    \multirow{2}{*}{Data normalization} & $\mu = \left(0.485, 0.456, 0.406\right)$ \\ & $\sigma = \left( 0.229, 0.224, 0.225 \right)$ \\
    \midrule
    \multicolumn{2}{c}{\textbf{Shrink \& Perturb}} \\
    \midrule
    Shrink $\shrink$ (default unless explicitly defined otherwise) & 0.4 \\
    Perturb $\perturb$ (default unless explicitly defined otherwise) & 0.1 \\
    \midrule
    \multicolumn{2}{ c }{\textbf{Layer-wise Re-initialization}} \\
    \midrule
    $\left(\numblocks, \lwriters\right)$ & $\left\lbrace \left(2, 1\right), \left(5, 1\right),\left(5, 2\right) 
    \right\rbrace$ \\
    \bottomrule
    \end{tabular}
    \caption{Hyperparameters common to all models trained on CIFAR-10 and CIFAR-100 ($32 \times 32$ images) \cite{cifar}.}
    \label{tbl:hyperparams-cifar}
\end{table*}

\begin{table*}[h]
    \centering
    \begin{tabular}{l c}
    \toprule
    \multicolumn{2}{c}{\textbf{General}}\\
    \midrule
    Total training epochs & 500 \\
    Stages $\times$ epochs per stage & $\left\lbrace 2 \times 100, 5 \times 40, 10 \times 20, 20 \times 10, 25 \times 8\right\rbrace$ \\
    Batch size     &  $100$ \\
    Momentum     & $0.9$ \\
    Learning rate grid & $\left\lbrace 0.01, 0.05, 0.1 \right\rbrace$ \\
    Weight decay grid & $\left\lbrace 0, 0.0001, 0.0005, 0.005 \right \rbrace$ \\
    Distillation strength $\distillcoeff$ (if used) & $2$ \\
    \multirow{2}{*}{Data augmentation (if used)} & Random horizontal flips and\\ & size-preserving crops after padding 4 pixels. \\
    \multirow{2}{*}{Data normalization} & $\mu = \left(0.485, 0.456, 0.406\right)$ \\ & $\sigma = \left( 0.229, 0.224, 0.225 \right)$ \\
    \midrule
    \multicolumn{2}{c}{\textbf{Shrink \& Perturb}} \\
    \midrule
    Shrink $\shrink$ (default unless explicitly defined otherwise) & 0.4 \\
    Perturb $\perturb$ (default unless explicitly defined otherwise) & 0.1 \\
    \bottomrule
    \end{tabular}
    \caption{Hyperparameters common to all models trained on Tiny ImageNet ($64 \times 64$ images) \cite{tiny}.}
    \label{tbl:hyperparams-tiny}
\end{table*}

\end{document}